\documentclass[10pt,twocolumn,letterpaper]{article}

\usepackage{cvpr}              % To produce the CAMERA-READY version
\usepackage{graphicx}
\usepackage{amsmath}
\usepackage{amssymb}
\usepackage{booktabs}

\usepackage[pagebackref,breaklinks,colorlinks]{hyperref}

\usepackage[capitalize]{cleveref}
\crefname{section}{Sec.}{Secs.}
\Crefname{section}{Section}{Sections}
\Crefname{table}{Table}{Tables}
\crefname{table}{Tab.}{Tabs.}

\def\cvprPaperID{*****} % *** Enter the CVPR Paper ID here
\def\confName{CVPR}
\def\confYear{2022}

\begin{document}

\title{UniUD-FBK-UB-UniBZ Submission to the EPIC-Kitchens-100 Multi-Instance Retrieval Challenge 2022}

\author{Alex Falcon\\
Fondazione Bruno Kessler and University of Udine\\
{\tt\small afalcon@fbk.eu}
\and
Giuseppe Serra\\
University of Udine\\
{\tt\small giuseppe.serra@uniud.it}
\and
Sergio Escalera\\
University of Barcelona and Computer Vision Center\\
{\tt\small sergio@maia.ub.es}
\and
Oswald Lanz\\
Free University of Bozen-Bolzano\\
{\tt\small lanz@inf.unibz.it}
}
\maketitle

%%%%%%%%% ABSTRACT
\begin{abstract}
   This report presents the technical details of our submission to the EPIC-Kitchens-100 Multi-Instance Retrieval Challenge 2022. To participate in the challenge, we designed an ensemble consisting of different models trained with two recently developed relevance-augmented versions of the widely used triplet loss. Our submission, visible on the public leaderboard, obtains an average score of 61.02\% nDCG and 49.77\% mAP.
\end{abstract}

%%%%%%%%% BODY TEXT
\section{Introduction\label{sec:intro}}
Retrieving the most relevant videos based on a user query is a difficult task involving joint visual and textual understanding. The EPIC-Kitchens-100 dataset \cite{Damen2021RESCALING} offers a challenging benchmark, comprising more than 70k egocentric video clips capturing activities from 45 kitchens. Differently from standard benchmarks in text-video retrieval, the EPIC-Kitchens-100 Multi-Instance Retrieval Challenge uses rank-aware metrics, such as the nDCG and the mAP, to assess the quality of the solutions. This is made possible by the introduction of a relevance function \cite{Damen2021RESCALING} which is defined in terms of the noun and verb classes found within the captions.

To participate in the challenge, we designed an ensemble of multiple models trained with two relevance-augmented versions \cite{falcon2022learning,falcon2022relevance} of the standard triplet loss \cite{schroff2015facenet}. The final results show the effectiveness of the training techniques we recently proposed, as well as the ensemble version. In particular, when compared to the public leaderboard from last year, we observe improvements of almost 8\% in nDCG and 5\% in mAP. Moreover, when compared to the current public leaderboard, we obtain the best result in terms of mAP with a margin of more than 2\%, and the second best result in nDCG with only 0.4\% difference.

In Section \ref{sec:strategies} we provide details about the two optimization strategies \cite{falcon2022learning,falcon2022relevance} which we recently developed. In Section \ref{sec:models} we describe the two architectures \cite{wray2019fine,chen2020fine} which we used as the basis of our study. Implementation details and a brief overview describing how we ensemble the different models are provided in Section \ref{sec:exp}. Finally, we conclude the report in Section \ref{sec:conclusion}.

\section{Optimization strategies\label{sec:strategies}}
We describe the details concerning two different optimization strategies which use the relevance function introduced in \cite{Damen2021RESCALING} to improve the contrastive loss functions commonly used to learn text-video retrieval models.

\subsection{Relevance-Margin}
To train a text-video retrieval model with the triplet loss function, the same fixed margin is enforced on the similarity between the anchor-positive pair and the anchor-negative pair. This strategy makes it possible to maximize the similarity of the descriptors of the video and caption pairs in the dataset. Yet, the negative examples may have different relevance values when compared to the anchor, and in particular they may be even partially relevant. Therefore, in \cite{falcon2022relevance} we proposed to replace the fixed margin with a relevance-based margin, that is a margin which is proportional to the relevance value of the video and caption descriptors which are to be contrasted. Given the anchor $a$, the positive $p$, and the negative $n$, it is defined as follows:
\begin{equation}
    \Delta_{a,p,n} = 1 - \mathcal{R}(a, n)
\end{equation}

\subsection{RANP}
Due to the sampling mechanisms used to form the triplets, all the negative examples are treated as equally irrelevant when compared to the anchor. Yet, as mentioned before, not all the negatives are actually irrelevant, and therefore those which are not completely irrelevant should not be treated as if they were. In \cite{falcon2022learning} we proposed RANP, a strategy which uses the relevance function and a threshold $\tau$ to separate relevant from irrelevant samples (up to a degree $\tau$) within the batch. By doing so, the negatives can be picked from a smaller negatives' pool which only contains irrelevant samples. Similarly, we introduced an additional triplet loss term which increases the similarity of the anchor with dissimilar yet relevant samples in the current network state during training. 

\section{Models\label{sec:models}}
We briefly describe here the two network architectures which we used as the base models.

\subsection{JPoSE} To have a fine-grained understanding of the actions in a retrieval setting, Wray et al. \cite{wray2019fine} introduced JPoSE, which disentangles the Part-of-Speech (PoS) in the captions in order to learn a multi-modal embedding space for each PoS tag. The PoS-restricted embeddings are then used to perform action retrieval in a joint embedding space. All the embedding spaces are finally learned by using a mix of PoS-restricted and PoS-agnostic losses.

\subsection{HGR} Chen et al. \cite{chen2020fine} propose to deal with fine-grained retrieval by means of hierarchical structures and graph reasoning. First of all, for each natural language description they build a graph of the semantic roles occurring between each noun and the associated verb phrase \cite{shi2019simple}. A global-to-local graph is then built by using these textual features as the nodes, which are then aggregated through graph message passing and aligned to the visual features with a bidirectional global loss term.

\section{Experiments\label{sec:exp}}
In this section, we detail the experimental settings of the models considered within the ensemble and the description of the ensembling strategy.

\subsection{Implementation details}
\noindent\textbf{Details of the models.} We briefly point out for each model some technical aspects related to the implementation details.
\begin{itemize}
    \item {\em Model 1.} JPoSE trained with the relevance-margin.
    \item {\em Model 2.} HGR trained with RANP, using $\tau=0.15$, $\Delta_p=0.2$ (see \cite{falcon2022learning} for more details about the margin $\Delta_p$).
    \item {\em Model 3.} HGR trained with RANP, using $\tau=0.15$, $\Delta_p=0.2$ with a lower size for the embedding space (512).
    \item {\em Model 4.} HGR trained with RANP, using $\tau=0.4$, $\Delta_p=0.25$.
    \item {\em Model 5.} HGR trained with RANP, using $\tau=0.4$, $\Delta_p=0.15$
\end{itemize}

\noindent\textbf{Training details.} We trained JPoSE by using the relevance-based margin within each of the triplet loss terms used in the method, including both cross-modal and within-modality losses, both at the global and at the PoS level. When dealing with the losses at the noun (respectively, verb) level, we set the verb IoU (respectively, noun IoU) to 1 during the computation of the relevance. The optimizer used is SGD with a momentum of 0.9 and learning rate 0.01. The model was trained for 100 epochs with a batch size of 64.

In the case of HGR, we used RANP as the training loss function. We employed Adam as the optimizer with a learning rate of 0.0001. We trained the model for 50 epochs with a batch size of 64.

\noindent\textbf{Dataset.} We used the full training set to train the models and we used a small validation set taken from the training set to keep track of the learning. We used the RGB, flow, and audio features extracted with TBN \cite{kazakos2019epic} which were provided by the dataset authors.

\subsection{Ensembling strategy}
After learning the aforementioned models, we created the similarity matrix of each of the five models. These are then summed before taking the mean similarity values. By doing so, the similarity of video $v_i$ and caption $q_j$ is computed as the mean of the five similarity values predicted by the models. Finally, we use this mean similarity matrix in the submission.

\subsection{Results}
In Table \ref{tab:results} we report the performance obtained by the various models used within the ensemble on the validation set. The final model which we submitted to the leaderboard (\textbf{Ens.}) obtained the best results on the validation set. Moreover, when compared to previous state-of-the-art approaches (from the EPIC-Kitchens-100 Multi-Instance Retrieval Challenge 2021 \cite{Damen2021CHALLENGES}), our ensemble shows considerable improvements: in fact, both Wray et al. (JPoSE trained without the relevance-based margin) and Hao et al. \cite{Damen2021CHALLENGES} obtained on average around 53\% nDCG and 44\% mAP, whereas our ensemble obtains around 61\% nDCG and almost 50\% mAP. On the other hand, when comparing to the current public leaderboard, we achieve top-1 mAP performance (49.77\% compared to 47.39\% obtained by the second best) and top-2 nDCG (61.02\% compared to 61.44\%).
\begin{table}[!t]
    \centering
    \begin{tabular}{c|ccc|ccc} \hline \hline
         & \multicolumn{6}{c}{Validation} \\ \hline
         & \multicolumn{3}{c|}{nDCG (\%)} & \multicolumn{3}{c}{mAP (\%)} \\ \hline % & \multicolumn{2}{c}{Local test (nDCG, mAP)} \\ \hline
        Mod. & v2t & v2t & avg & v2t & v2t & avg \\ \hline % & avg & avg \\ \hline
        1 & 74.6 & 71.1 & 72.8 & 78.7 & 74.1 & 76.4 \\ \hline %  & 56.2 & 45.8 \\
        2 & 81.2 & 77.4 & 79.3 & 85.4 & 75.4 & 80.4 \\ \hline %  & 58.8 & 47.2 \\
        3 & 81.4 & 77.5 & 79.5 & 85.0 & 74.3 & 79.7 \\ \hline %  & 58.8 & 46.8 \\
        4 & 81.7 & 77.7 & 79.7 & 85.0 & 72.6 & 78.8 \\ \hline %  & 59.4 & 46.2 \\
        5 & 82.0 & 78.1 & 80.1 & 86.4 & 75.8 & 81.1 \\ \hline %  & 59.2 & 46.8 \\
        \textbf{Ens.} & \textbf{82.8} & \textbf{79.5} & \textbf{81.2} & \textbf{88.2} & \textbf{78.7} & \textbf{83.5}  \\ \hline % & 61.0 & 49.8 \\ \hline
         & \multicolumn{6}{c}{Official test} \\ \hline
         & \multicolumn{3}{c|}{nDCG (\%)} & \multicolumn{3}{c}{mAP (\%)} \\ \hline %  & \multicolumn{2}{c}{} \\ \hline
        \textbf{Ens.} & \textbf{63.16} & \textbf{58.88} & \textbf{61.02} & \textbf{55.15} & \textbf{44.39} & \textbf{49.77}  \\ \hline
        %JPoSE & 55.51 & 51.55 & 53.53 & 49.91 & 38.11 & 44.01 \\
        %Hao & 55.28 & 51.83 & 53.56 & 49.96 & 38.49 & 44.23 \\ \hline
        \hline % & - & - \\ \hline \hline
    \end{tabular}
    \caption{Performance of the five considered models on the validation set (top) and test set (bottom) of EPIC-Kitchens-100. The similarity scores predicted by the ensemble are obtained by averaging the predictions made by each individual model. }%We also added the two best results (Hao et al. \cite{Damen2021CHALLENGES} and JPoSE) from EPIC-Kitchens-100 Multi-Instance Retrieval Challenge 2021.}
    \label{tab:results}
\end{table}

\section{Conclusion\label{sec:conclusion}} In this report, we summarized the details of our submission to the EPIC-Kitchens-100 Multi-Instance Retrieval Challenge 2022. The proposed ensemble, comprising several models trained with relevance-augmented version of the standard triplet loss, achieves considerable improvements when compared to last year challenge competitors. Moreover, the result we obtain is visible on the public leaderboard and obtains top-1 performance in mAP (with a margin of 2.4\%) and top-2 performance in nDCG (with a difference of 0.4\%).

\section*{Acknowledgements} We gratefully acknowledge the support from Amazon AWS Machine Learning Research Awards (MLRA) and NVIDIA AI Technology Centre (NVAITC), EMEA. We acknowledge the CINECA award under the ISCRA initiative, which provided computing resources for this work.

%%%%%%%%% REFERENCES
{\small
\bibliographystyle{ieee_fullname}
\bibliography{report}
}

\end{document}